\documentclass[conference, twocolumn]{IEEEtran}

\counterwithin*{paragraph}{subsection}

\usepackage{listings}
\usepackage{graphicx}
\usepackage{subcaption}
\usepackage[table,xcdraw]{xcolor}
\usepackage{amsmath}
\usepackage{amssymb}
\usepackage{pifont}
\definecolor{mydarkblue}{rgb}{0,0.08,0.65}
\usepackage[colorlinks=true,
    linkcolor=mydarkblue,
    citecolor=mydarkblue,
    filecolor=mydarkblue,
    urlcolor=mydarkblue]{hyperref}
\hypersetup{
  pdftitle={Expert Coupling in MoE Pretraining: Reducing All-to-All Overhead with Correlated Placement and Token Shuffling},
  pdfauthor={Radha Gulhane, Quentin Anthony, Beren Millidge}
}
\usepackage{cleveref}
\usepackage{soul}
\usepackage[shortlabels]{enumitem}
\usepackage{url}
\usepackage{color}
\usepackage{tikz}
\usepackage{balance}
\usepackage{multirow}
\usepackage{comment}
\usepackage{wrapfig,lipsum,booktabs}
\usepackage[frozencache,cachedir=.]{minted}
\usepackage[symbol]{footmisc}
\usepackage{tablefootnote}
\usepackage{tabularx}
\usepackage{placeins}
\usepackage{algorithm}
\usepackage{algpseudocode}
\usepackage{cuted}
\usepackage{float}
\usepackage[compatibility=false]{caption}
\usepackage{multicol}
\usepackage{dblfloatfix}
\usepackage{natbib}
\usepackage{algorithm}
\usepackage{algpseudocode}
\usepackage{amsmath,amssymb}
\usepackage{booktabs}

\usepackage{xcolor}
\usepackage{caption}

\usepackage{amsmath}
\usepackage{tikz}
\usepackage{placeins}   
\usepackage{booktabs}
\usepackage[table]{xcolor}
\usepackage{multirow}

\newcommand*\stepnum[1]{%
  \tikz[baseline=(n.base)]\node[
    circle,
    fill=gray!25,        
    text=black!75,       
    inner sep=0pt,
    minimum size=1.15em,
    font=\scriptsize\sffamily
  ] (n) {#1};%
}
\renewcommand{\arraystretch}{1.2}
\newcommand{\corr}{\textsuperscript{*}}

\definecolor{codegreen}{rgb}{0,0.6,0}
\definecolor{codegray}{rgb}{0.5,0.5,0.5}
\definecolor{codepurple}{rgb}{0.58,0,0.82}
\definecolor{backcolour}{rgb}{0.95,0.95,0.92}
\lstdefinestyle{mystyle}{
  backgroundcolor=\color{backcolour},
  commentstyle=\color{codegreen},
  keywordstyle=\color{magenta},
  numberstyle=\tiny\color{codegray},
  stringstyle=\color{codepurple},
  basicstyle=\ttfamily\footnotesize,
  breakatwhitespace=false,
  breaklines=true,
  captionpos=b,
  keepspaces=true,
  numbers=left,
  numbersep=5pt,
  showspaces=false,
  showstringspaces=false,
  showtabs=false,
  tabsize=2,
}
\makeatletter
\def\blfootnote{\xdef\@thefnmark{}\@footnotetext}
\makeatother
\AtBeginDocument{%
  \providecommand\BibTeX{{%
    \normalfont B\kern-0.5em{\scshape i\kern-0.25em b}\kern-0.8em\TeX}}}

\IEEEoverridecommandlockouts

\begin{document}

\title{\LARGE Expert Coupling in MoE Pretraining: Reducing All-to-All Overhead with Correlated Placement and Token Shuffling}
%
%


\author{
Radha Gulhane\corr $\quad$
Quentin Anthony $\quad$
Beren Millidge $\quad$ \\ \\
{\textbf{Zyphra}}
}

\maketitle

\blfootnote{\textsuperscript{*}Corresponding author: \texttt{radha@zyphra.com}}

\begin{abstract}
Mixture-of-Experts (MoE) layers replace the feed-forward block of a Transformer with $E$ expert networks, and each token is routed to $k$ of these experts. Under expert parallelism (EP) the experts are distributed across GPUs, and every MoE layer runs all-to-all collectives in the forward and backward passes to dispatch tokens to their experts and then combine the results. On a cluster with 8 AMD Instinct MI300X GPUs per node, these collectives can take 45\,\% of the training step at EP32 with top-2 routing and 60\,\% with top-6 routing. We find that early in pretraining routers have already learned to assign tokens to experts in correlated patterns, both within a layer and across layers. At top-2, 0.8\,\% of the expert pairs in a layer are selected together by 42\,\% of tokens, and the experts a token selects at one layer predict the experts it selects at the next layer. We use these correlations to keep more token--expert assignments on the token's own GPU, which reduces communication across GPUs and across nodes. Correlated expert placement puts experts that are often selected together on the same GPU. Combined with a dispatcher that sends each token to each GPU once, it removes up to 58\,\% of dispatched rows. Token shuffling applies when sequence parallelism shards tokens across the EP group. It moves each token to the GPU predicted to hold its next-layer experts during the reduce-scatter that follows attention. On one node this raises the share of token--expert assignments served on the token's GPU from 12.5\,\% to 59\,\%. In Megatron-LM, across EP degrees from 8 to 64 with top-2 and top-6 routing, the two methods reduce all-to-all time by 1.16--2.63$\times$ and end-to-end step time by up to 1.41$\times$. Neither method changes the models' underlying routing decisions or expert parameters.
\end{abstract}


\section{Introduction}
Mixture-of-Experts (MoE) layers have become a central building block of large language models~\citep{mixtral,deepseekv3,qwen3,kimik3}. In an MoE layer, the single feed-forward network (FFN) of a dense Transformer block is replaced by a set of $E$ parallel FFNs, called \emph{experts}, together with a lightweight router~\citep{shazeer2017moe,gshard,switch}. After the attention sub-layer, the router scores each token's hidden state against all experts and selects its top-$k$ experts. The token is processed only by those experts, and their outputs are combined using the router weights. Model capacity therefore grows with $E$, while the FLOPs per token grow only with $k$. Because each expert is updated only by the tokens routed to it, experts become specialized over the course of pretraining~\citep{deepseekmoe,openmoe}.

This capacity comes at a communication cost. Under expert parallelism (EP)~\citep{gshard}, experts are sharded across GPUs, and every MoE layer requires two all-to-all (A2A) collectives. The \emph{dispatch} sends each token from the GPU that owns it to the GPUs holding its selected experts, and the \emph{combine} returns the expert outputs to the owning GPU. Both collectives run in the forward pass and again in the backward pass. The communication volume grows with the EP degree and with top-$k$, and once the EP group spans multiple nodes, the slower inter-node links dominate step time. In our measurements on a cluster with 8 AMD Instinct MI300X GPUs per node (Section~\ref{sec:mot-a2a}), all-to-all communication accounts for 13.5\,\% and 44.8\,\% of step time at EP8 and EP32 under top-2 routing, and for 24.0\,\% and 60.4\,\% under top-6 routing.

Conventional MoE implementations place experts contiguously across the EP group, with expert $e$ on GPU $\lfloor e/(E/\mathrm{EP}) \rfloor$, irrespective of which experts are used together. Depending on the routing, the all-to-all can also become unbalanced, and the few GPUs that receive a disproportionate share of tokens become stragglers that delay the whole collective. Mechanisms that address this include expert replication~\citep{deepseekv3}, load-driven expert relocation~\citep{smartmoe}, and dynamically planned redundant experts with weight prefetching~\citep{moonep,kimik3}. SmartMoE also changes which GPU holds each expert, but it places experts to balance load, not to co-locate experts that are selected together. These techniques lower the \emph{peak} per-link traffic by distributing the traffic across GPUs. The \emph{total} traffic stays about the same, because the same fraction of token--expert assignments crosses a GPU boundary. Replication also costs memory for the expert copies. DeepSeek-V3 additionally reduces inter-node traffic with node-limited routing, which restricts the experts the router can select.

A second line of work exploits expert \emph{affinity} for MoE inference. GRACE-MoE~\citep{gracemoe} groups experts that are often selected together within a layer onto the same GPU, first across nodes and then across the GPUs of each node, and sends a token to each GPU only once. Its groups differ in size, and it replicates heavily loaded experts to rebalance the GPUs, which costs memory. We apply within-layer grouping to pretraining with equal-size groups and no replicas. ExFlow~\citep{exflow} uses inter-layer expert affinity to co-locate experts across layers. This approach does not transfer directly to pretraining. During autoregressive decoding, a token can be placed on any convenient GPU. During pretraining, a token stays on the GPU that holds its part of the sequence for the attention sub-layer, and moving it requires additional communication. Exploiting cross-layer affinity during training therefore requires a mechanism that ExFlow does not provide.

Our starting point is that once experts specialize during pretraining, routing follows stable patterns. We study expert routing across pretraining stages and load-balancing methods, and find two such patterns. The first is \emph{within-layer correlation}. Some pairs of experts are selected together much more often than independent routing would predict. After approximately 2.1\,B training tokens with top-2 routing, 42\,\% of tokens select one of only 0.8\,\% of the possible expert pairs, and this pattern is already present after about one billion tokens. The second is \emph{cross-layer correlation}. The experts a token selects at layer $\ell$ predict the experts it selects at layer $\ell+1$. Both correlations appear with all three load-balancing methods we test, but with differing strengths.

We exploit these correlations to raise \emph{token--expert locality}, the fraction of (token, selected expert) pairs whose expert is on the token's GPU. These pairs never enter the network, so higher locality reduces the total A2A volume. Our methods require no architectural changes or expert replicas, and they do not change routing decisions or the training loss. Correlated placement and deduplication apply to any layout that uses expert parallelism. Token shuffling applies only to layouts that use tensor parallelism with sequence parallelism and in which the TP group and the EP group are the same set of GPUs ($\mathrm{TP}=\mathrm{EP}$), such as TP8~EP8 and TP16~EP16 in our evaluation. It does not apply to layouts without tensor parallelism (TP1). Correlated expert placement and deduplication also apply to inference, where affinity-based grouping has already been evaluated~\citep{gracemoe}; we do not evaluate inference.

We make the following contributions:

\begin{itemize}
  \item \textbf{Correlated Expert Placement and a Deduplicating Dispatcher (Section~\ref{sec:within}).} We place correlated experts on the same GPU, so that the experts a token selects are more likely to share a GPU. The deduplicating dispatcher sends each token to a destination GPU at most once, regardless of how many of its selected experts are there. Deduplication is an existing technique~\citep{deepep}. The all-to-all dispatcher in Megatron-LM~\citep{megatron-lm} sends one copy of a token per selected expert, so we implement deduplication there as the basis for placement. Together, placement and deduplication reduce both intra-node and inter-node communication volume.

  \item \textbf{Token Shuffling (Section~\ref{sec:shuffle}).} For sequence-parallel layouts in which the TP group equals the EP group, we (i) predict a token's experts at layer $\ell+1$ from the experts the token selected at the previous layers, and (ii) move the token ahead of time to the GPU that is predicted to hold those experts. The move is folded into the reduce-scatter that sequence parallelism~\citep{sequence_parallel} already performs after attention. This brings to pretraining the locality benefit that prior affinity-based work obtained for inference.

  \item \textbf{Characterization of Expert Routing (Sections~\ref{sec:affinity} and~\ref{sec:results}, and the appendix).} We quantify within-layer co-selection and cross-layer predictability across pretraining stages, top-$k$ settings and load-balancing methods, and relate these statistics to communication volume.

  \item \textbf{Evaluation in Megatron-LM (Section~\ref{sec:results}).} We report communication volume, A2A time and end-to-end step time. With expert parallelism alone, correlated placement and deduplication reduce A2A time by $1.16\times$--$1.38\times$ under top-2 routing and by $1.35\times$--$1.95\times$ under top-6 routing. With token shuffling added, A2A time is reduced by $1.40\times$--$1.74\times$ under top-2 and by $1.46\times$--$2.63\times$ under top-6.
\end{itemize}


\section{Motivation}
\FloatBarrier 
\label{sec:motivation}

\subsection{All-to-All Cost in MoE Pretraining}
\label{sec:mot-a2a}

\begin{figure*}[!ht]
\centering
\includegraphics[width=0.7\linewidth]{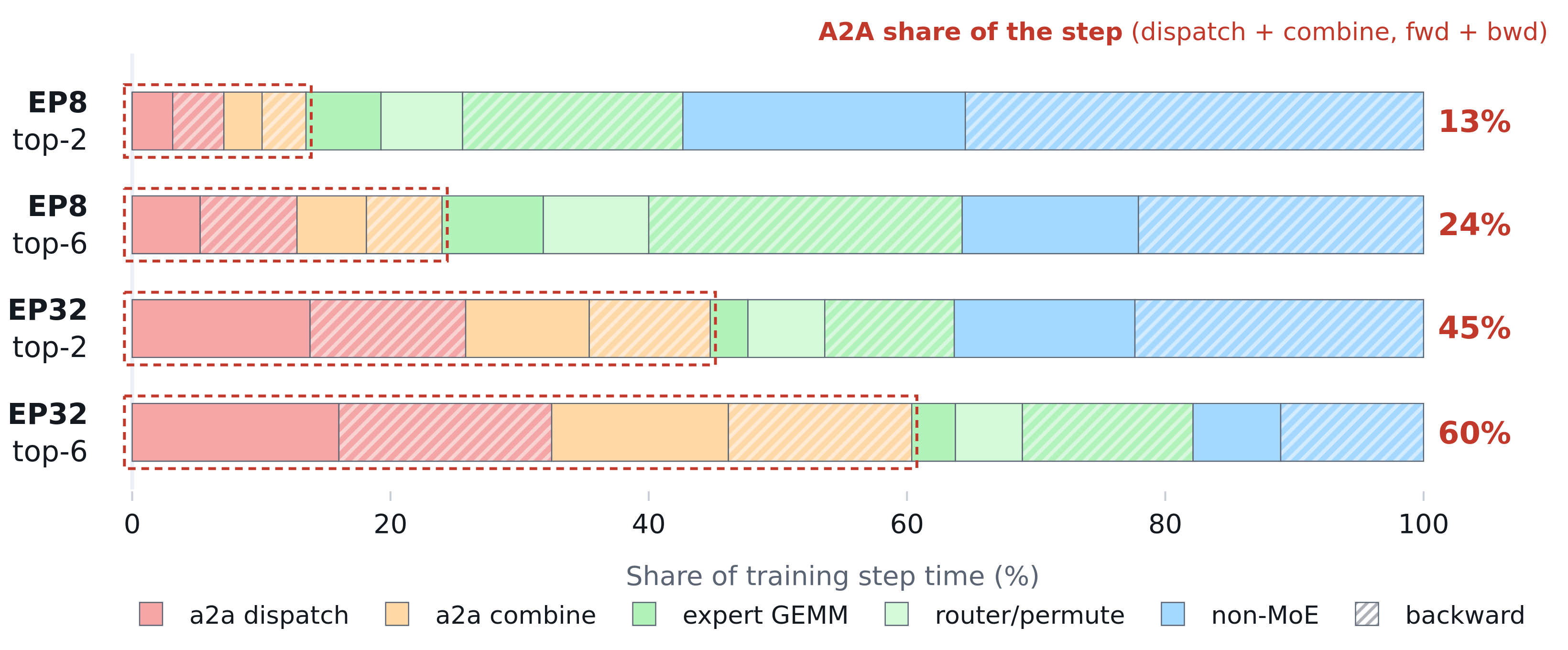}
\caption{Share of training step time by component at EP8 and EP32 with top-2 and top-6 routing, using the Megatron-LM all-to-all dispatcher and contiguous placement. Dashed boxes mark the all-to-all share.}
\label{fig:step_anatomy_P2_TP1_top2-6_EP8-32}
\end{figure*}

We begin by measuring how much of an MoE training step is spent in the dispatch and combine collectives. Each MoE layer runs both collectives in the forward pass and again in the backward pass. The cost of these collectives depends on two configuration choices. A larger expert-parallel degree spreads experts over more GPUs, so the fraction of a token's experts on its own GPU falls as $1/\mathrm{EP}$. Beyond a single node, however, the exchange also moves from the intra-node fabric to the inter-node network. A larger top-$k$ raises the number of activation rows each token contributes to both collectives, so top-6 moves three times the volume of top-2.

We measure a 12-layer MoE model ($d_\mathrm{model}{=}4096$), with 128 experts of FFN width 1024 per layer (see Section~\ref{sec:results} and Table~\ref{tab:p2_arch} for the detailed configuration). Figure~\ref{fig:step_anatomy_P2_TP1_top2-6_EP8-32} shows that the two effects compound. We measure all-to-all time as the wall-clock time the step spends inside the dispatch and combine calls, which includes time spent waiting for other GPUs. Widening the EP group from a single node (EP8) to four nodes (EP32) raises the all-to-all share of step time from 13.5\,\% to 44.8\,\% under top-2 and from 24.0\,\% to 60.4\,\% under top-6. At fixed EP8, moving from top-2 to top-6 raises it from 13.5\,\% to 24.0\,\%. At EP32 with top-6, the two collectives take more of the step than expert computation and all non-MoE computation combined.

\subsection{Expert Correlation Within and Across Layers}
\label{sec:affinity}

Crucially, the measurements above use contiguous placement, which ignores the correlations between the experts a token selects. Once experts have specialized, these selections follow two regular patterns, which we exploit in this paper. We measure both on the top-2 and top-6 models from Section~\ref{sec:mot-a2a}. All statistics in this section are computed offline. We record every token's expert choices at every MoE layer on a few late training steps (131k tokens per step), fit each placement and predictor on some of these steps, and apply them to the recorded routes of the other steps without running the dispatcher.
\begin{figure}[!ht]
  \centering
  \captionsetup[sub]{font=footnotesize}
  \begin{subfigure}[t]{0.5\columnwidth}
    \centering
    \includegraphics[width=\linewidth]{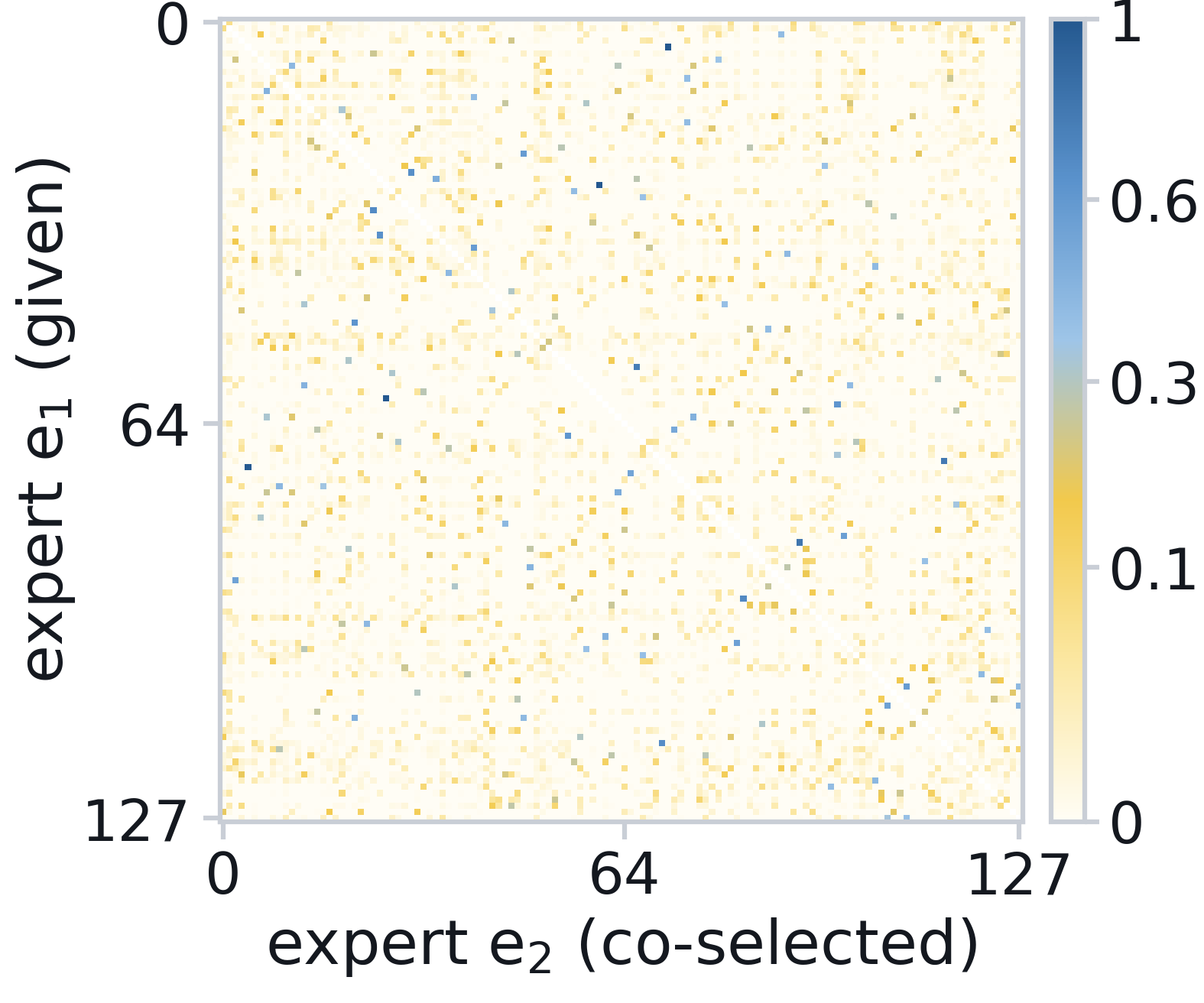}
    \caption{Within layer 8}
    \label{fig:within-affinity}
  \end{subfigure}\hfill
  \begin{subfigure}[t]{0.5\columnwidth}
    \centering
    \includegraphics[width=\linewidth]{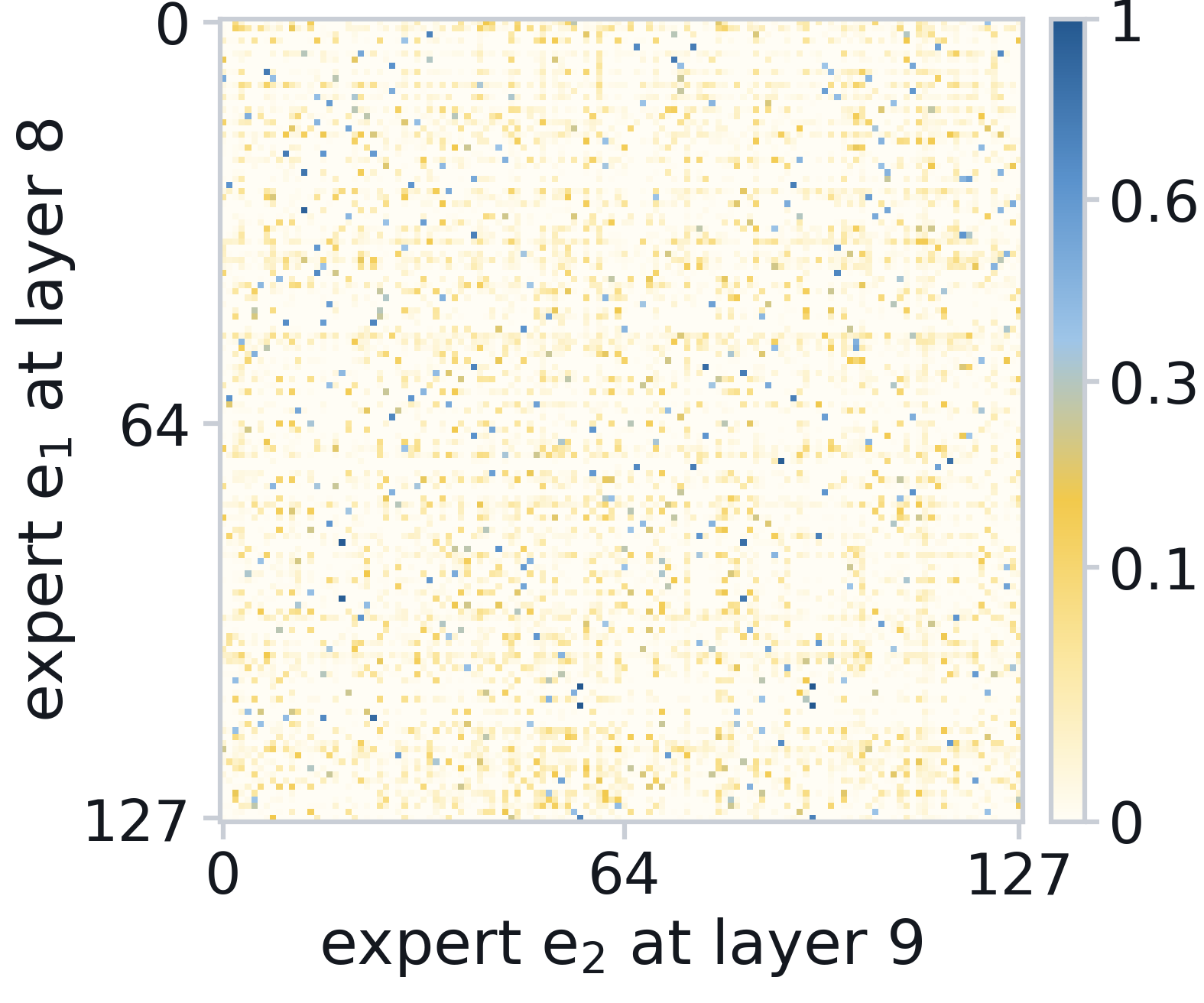}
    \caption{Across layers $8\to9$}
    \label{fig:across-affinity}
  \end{subfigure}
  \caption{Expert correlation in the top-2 model, computed from the saved routing of 524k tokens at training steps 1983--1995 (every fourth step). (a)~$P(e_2 \in \text{top-2} \mid e_1 \in \text{top-2})$ within layer 8; (b)~$P(e_2 \text{ at } 9 \mid e_1 \text{ at } 8)$ across layers. Blue marks strongly correlated pairs (conditional probability above 0.3).}
  \label{fig:affinity-heatmaps}
\end{figure}
\begin{table}[!ht]
\centering
\caption{Offline placement and prediction statistics on held-out iterations
1983 and 1987 (fit on 1991--1999); all values in \% (rounded) except
\emph{GPUs}. \emph{Same GPU}: tokens whose top-$k$ experts share a GPU.
\emph{GPUs}: distinct destinations per token. \emph{Deduplication}: rows
removed as duplicates. \emph{Local experts}: share of a token's next-layer
experts on its own GPU when it is moved to the GPU predicted from its
current experts ($T/\mathrm{EP}$ tokens per GPU) and when it stays on its
home GPU ($1/\mathrm{EP}$).}
\label{tab:placement}
\setlength{\tabcolsep}{4pt}
\renewcommand{\arraystretch}{1.05}
\begin{tabular}{@{}llccccc@{}}
\toprule
& & \multicolumn{3}{c}{\textbf{Within layer}} & \multicolumn{2}{c}{\textbf{\shortstack[c]{Across layers:\\local experts}}} \\
\cmidrule(lr){3-5}\cmidrule(l){6-7}
\textbf{EP} & \textbf{Placement}
  & \textbf{\shortstack[c]{Same\\GPU}}
  & \textbf{GPUs}
  & \textbf{\shortstack[c]{Dedupli-\\cation}}
  & \textbf{\shortstack[c]{Token\\moved}}
  & \textbf{\shortstack[c]{Token\\stays}} \\
\midrule
\rowcolor{gray!12}
\multicolumn{7}{c}{\textbf{\textit{top-2 routing}}} \\
\multirow{2}{*}{EP8}  & Contiguous & 12 & 1.88 &  6 & 33 & 13 \\
                      & Correlated   & 53 & 1.47 & 26 & 58 & 13 \\
\cmidrule(lr){1-7}
\multirow{2}{*}{EP16} & Contiguous &  6 & 1.95 &  3 & 28 &  6 \\
                      & Correlated   & 45 & 1.55 & 23 & 51 &  6 \\
\cmidrule(lr){1-7}
\multirow{2}{*}{EP32} & Contiguous &  2 & 1.98 &  1 & 25 &  3 \\
                      & Correlated   & 36 & 1.64 & 18 & 42 &  3 \\
\midrule
\rowcolor{gray!12}
\multicolumn{7}{c}{\textbf{\textit{top-6 routing}}} \\
\multirow{2}{*}{EP8}  & Contiguous &  0 & 4.49 & 25 & 25 & 13 \\
                      & Correlated   & 18 & 2.54 & 58 & 60 & 13 \\
\cmidrule(lr){1-7}
\multirow{2}{*}{EP16} & Contiguous &  0 & 5.24 & 13 & 18 &  6 \\
                      & Correlated   &  8 & 3.08 & 49 & 49 &  6 \\
\cmidrule(lr){1-7}
\multirow{2}{*}{EP32} & Contiguous &  0 & 5.67 &  6 & 14 &  3 \\
                      & Correlated   &  0 & 3.89 & 35 & 33 &  3 \\
\bottomrule
\end{tabular}
\end{table}

\paragraph{Within a Layer, Experts Are Selected in Stable Pairs}\label{par:within}
Figure~\ref{fig:within-affinity} shows, for layer~8, the probability
that a token selecting expert $e_1$ also selects $e_2$. The
blue cells are pairs that are often chosen together: 14 pairs are chosen
with at least 50\,\% confidence and carry 16\,\% of all tokens, and 64 of
the 8{,}128 possible pairs carry 42\,\%, whereas independent routing with
the same loads would give 1.6\,\%. For 73 of the 128 experts, one partner
appears in at least 30\,\% of its tokens.

A token is sent once per GPU its experts occupy, so co-locating a
frequent pair halves its traffic under a deduplicating dispatcher.
Contiguous placement does this only by chance; correlated placement does it on purpose.
Table~\ref{tab:placement} (within layer) counts rows, with each row corresponding to one copy of
a token's hidden state sent to one GPU. At EP8, correlated placement
raises the share of top-2 tokens whose experts share a GPU from 12\,\% to
53\,\%, and raises the fraction of rows removed by deduplication from 6\,\% to 26\,\% under
top-2 and 25\,\% to 58\,\% under top-6. At EP32, a GPU holds only four
experts, so a top-6 token's experts can never all share one, yet rows
removed still rise from 5.5\,\% to 35.1\,\%.

\paragraph{Across Layers, the Current Experts Predict the Next Ones}
\label{par:across-affinity}

Figure~\ref{fig:across-affinity} plots, for each expert $e_1$ at
layer~8, the probability that a token routed to $e_1$ selects expert
$e_2$ at layer~9. Most layer-8 experts send a large share of their tokens
to a single layer-9 expert: 112 of the 128 send at least 30\,\% to one
specific expert, and the median share is 48\,\%.

This correlation can be used to move \emph{tokens}. Before the router of
layer $i{+}1$ runs, the experts chosen at layer $i$ already indicate
which GPU will hold most next experts, and under sequence
parallelism (Section~\ref{sec:prelim-tpep}) the token can be moved there during attention. Every GPU must own exactly $T/\mathrm{EP}$
tokens, so not every token gets its predicted GPU.
Table~\ref{tab:placement} (across layers) reports, under this
constraint, the share of a token's next-layer experts that are local. A token that stays on its home GPU has $1/\mathrm{EP}$ local. A token moved to its predicted GPU has 58\,\%, 51\,\% and 42\,\%
at EP8, EP16 and EP32 for top-2 and 60\,\%, 49\,\% and 33\,\% for top-6
under correlated placement. Placement matters here too: with contiguous placement the moved token has only 33\,\% of its experts local at EP8, because its next experts span more GPUs.


\section{Background}
\begin{figure*}[!t]
    \centering
    \includegraphics[width=\textwidth]{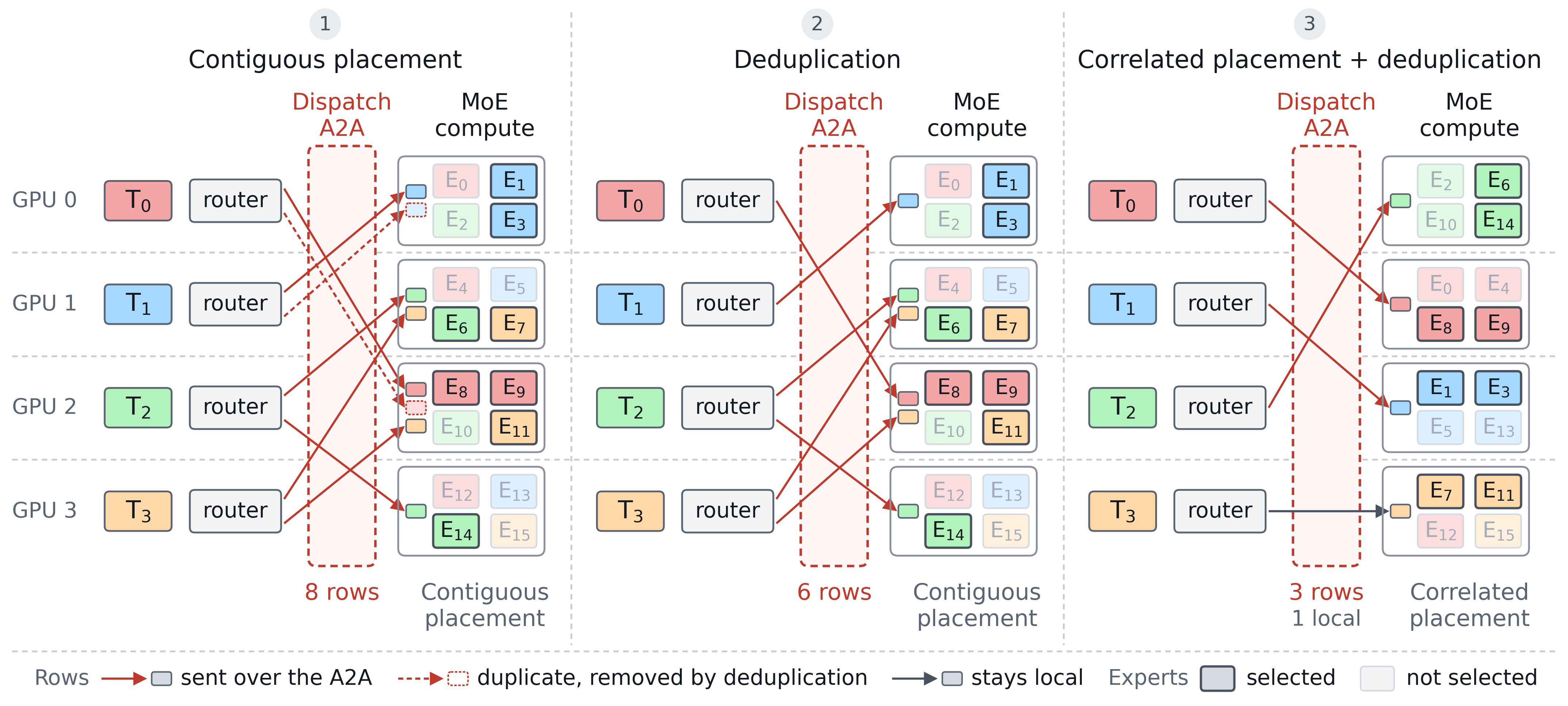}
    \caption{Reducing A2A with within-layer expert correlation. Each token $T_i$ and its selected experts have the same color.}
    \label{fig:coactivation}
\end{figure*}

\subsection{Expert Parallelism}
\label{sec:prelim-ep}

When the experts of a layer no longer fit in the memory of one GPU, expert parallelism (EP) partitions them across a group of GPUs, typically alongside data parallelism. With $E{=}128$ experts and $\mathrm{EP}{=}8$, for example, each GPU hosts $E/\mathrm{EP}{=}16$ experts. A token routed to an expert on another GPU is sent there by the dispatch all-to-all and returned by the combine all-to-all, in both the forward and the backward pass. The first panel of Figure~\ref{fig:coactivation} shows this pattern. Every token--expert assignment that crosses a GPU boundary adds a row to both collectives.

\subsection{Hybrid Parallelism: Tensor--Expert Parallelism}
\label{sec:prelim-tpep}

For large models and long-context training, expert parallelism alone is not enough to fit the model in GPU memory, and additional model parallelism is needed. Tensor parallelism (TP)~\citep{megatron-lm} shards the computation of each layer across GPUs and is widely used in both pretraining and inference.

In a common hybrid TP--EP configuration, attention and the MoE layer use different token layouts. Attention is partitioned across the TP ranks, and the MoE layer operates on sequence shards distributed across the same ranks~\citep{sequence_parallel,moe_folding,hybrid_parallelism}. Two collectives move between these layouts. After attention, a reduce-scatter sums the partial TP outputs and partitions the sequence, which leaves each rank with a contiguous subset of tokens. These tokens then pass through the dispatch all-to-all, the expert computation, and the combine all-to-all. After the MoE layer, an all-gather reconstructs the full sequence on every TP rank for the next Transformer block.

Token shuffling replaces this reduce-scatter and all-gather with shuffled versions (Figure~\ref{fig:shuffled_tp_ep_e2e}), and it requires the TP group and the EP group to be the same GPUs ($\mathrm{TP}=\mathrm{EP}$). The shuffled reduce-scatter sends each token to the rank predicted to hold most of its experts in the coming MoE layer, based on the experts it selected in the previous layers. The shuffle-aware all-gather restores the canonical token order before the next attention layer. A larger fraction of each token's experts is then on its own GPU, and the dispatch and combine all-to-alls send fewer bytes across the network. Section~\ref{sec:shuffle} describes the mechanism in detail. Experts are not tensor-parallel in this work. Some latency-sensitive inference deployments shard experts with tensor parallelism, which we do not consider.

\section{Related Work}
\label{sec:related}
Two systems place experts by within-layer affinity on models that are already trained. Occult~\citep{occult} measures co-selection in a pretrained MoE model on a profiling dataset, places experts that are often selected together within a layer on the same GPU, and sends a token to each GPU only once. It evaluates inference and fine-tuning, and obtains further savings by restricting each token's experts to a few GPUs, which changes routing decisions. GRACE-MoE~\citep{gracemoe} targets inference. It groups co-selected experts first across nodes and then across the GPUs of each node, lets the groups differ in size, and replicates heavily loaded experts to rebalance the GPUs, which costs memory. Correlated expert placement (Section~\ref{sec:within}) uses the same within-layer signal and, like GRACE-MoE, partitions across nodes before GPUs. Our work differs in three ways. We study pretraining from scratch, where the correlations do not exist at the start, and measure how early they form and how few tokens are needed to estimate them (Section~\ref{sec:res-corr}). We leave routing decisions unchanged and add no replicas. We also move tokens to their experts with token shuffling (Section~\ref{sec:shuffle}), which neither system does.


\section{Implementation}
\label{sec:implementation}
\subsection{Within-Layer Correlation: Correlated Expert Placement}
\label{sec:within}
\label{sec:expert-placement}
As discussed in Section~\ref{sec:affinity} and shown in Figure~\ref{fig:within-affinity}, some experts within a layer are frequently selected together by the same tokens. We use this structure to reduce all-to-all (A2A) communication with two techniques. \emph{Correlated expert placement} puts frequently co-selected experts on the same GPU, and a \emph{deduplicating dispatcher} sends each token to each EP rank at most once.

Figure~\ref{fig:coactivation} shows the A2A communication for three configurations:  the baseline, with deduplication, and with both deduplication and correlated expert placement. \protect\stepnum{1} shows contiguous placement with the conventional dispatcher, where each $(\text{token},\text{expert})$ assignment produces a separate A2A row. When a token selects several experts on the same destination rank, its hidden state is sent there several times. \protect\stepnum{2} adds deduplication. Assignments that share the same $(\text{token},\text{destination rank})$ are merged, so the hidden state is sent to that rank once. \protect\stepnum{3} adds correlated expert placement, which puts frequently co-selected experts on the same rank. This increases token--expert locality and gives deduplication more rows to merge. Neither technique changes the router's expert assignments or the expert computation.

\paragraph{Deduplicating Dispatcher}
Conventional top-$k$ dispatch sends one copy of the token's hidden state per $(\text{token},\text{expert})$ assignment, even when several selected experts reside on the same EP rank. We instead send one row per unique $(\text{token},\text{destination rank})$ pair. Each row carries the IDs and routing probabilities of the token's experts on that rank. The receiving rank uses this metadata to give each of its selected experts a copy of the token and to group the copies by expert for the expert computation, so no separate collective is needed for this metadata. The combine path works the same way. The expert rank weights the outputs of the token's experts by their routing probabilities, sums them, and returns one partial output per $(\text{token},\text{rank})$ pair. Dispatch and combine therefore send one row per token for each rank that holds at least one of its experts. Deduplication is an existing technique~\citep{deepep}, and we implement it in Megatron-LM as the basis for our correlated placement algorithms.

\paragraph{Correlated Placement}
For each MoE layer, we build a correlation graph $W$, where $W[e,e']$ counts the tokens that select both experts $e$ and $e'$. We look for a placement that puts the same number of experts on every GPU and maximizes the total weight $W[e,e']$ of the expert pairs that share a GPU. Algorithm~\ref{alg:correlative-placement} in the appendix performs a Kernighan--Lin-style local search~\citep{kernighan1970}. It starts from contiguous placement and swaps pairs of experts between GPUs whenever a swap increases this total. Every GPU holds exactly $E/\mathrm{EP}$ experts, so the placement does not change the memory used for expert parameters.

For EP groups that span multiple nodes, we partition in two steps. We first partition the experts across nodes, which keeps co-selected experts on the same node and reduces inter-node traffic. We then partition each node's experts across its GPUs, which keeps co-selected experts on the same GPU where possible. The resulting placement is a per-layer permutation of global expert IDs and does not change router decisions or expert parameters.

\subsection{Cross-Layer Correlation: Token Shuffling}
\label{sec:shuffle}

\begin{figure}[!h]
\centering
\includegraphics[width=1\columnwidth]{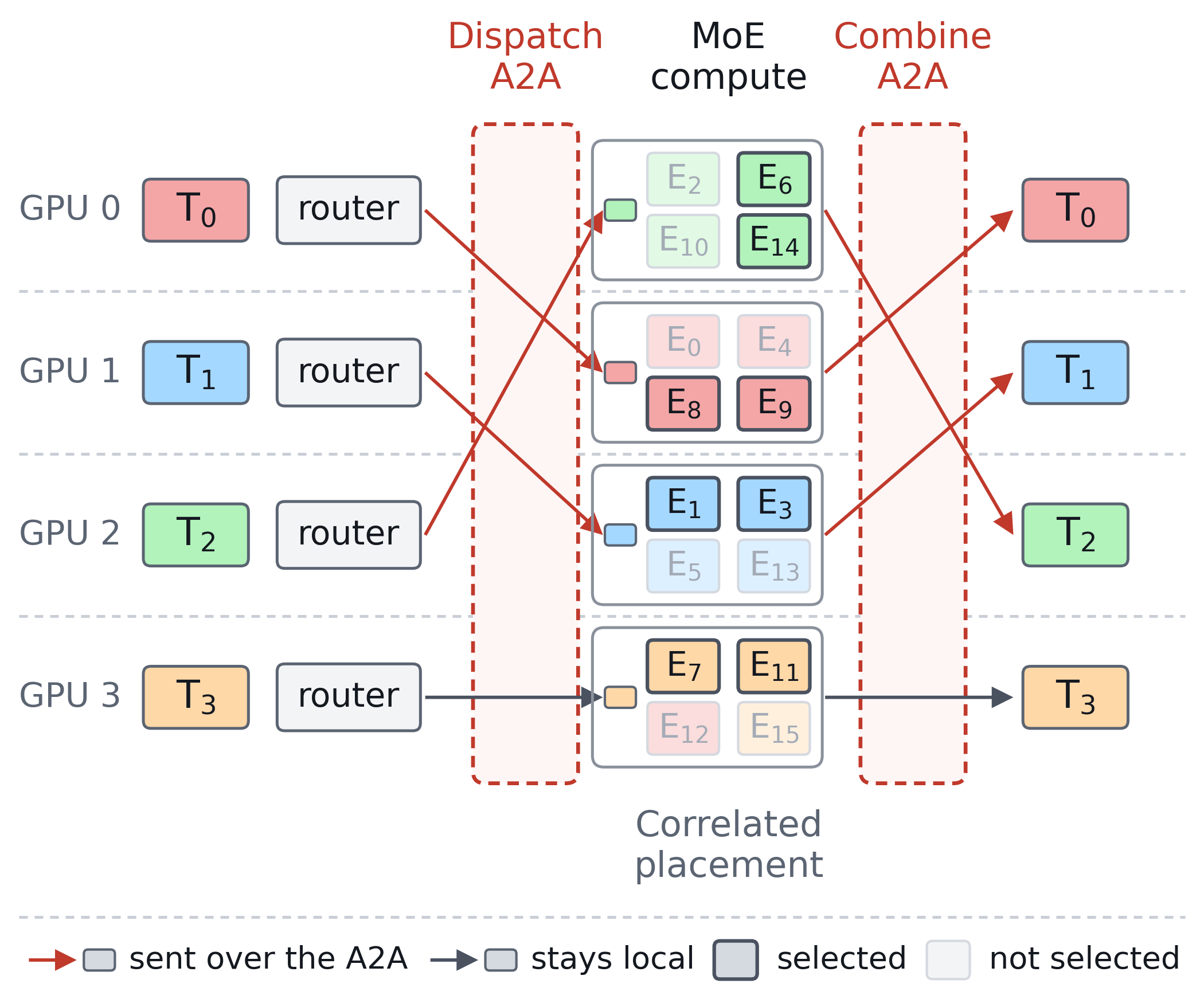}
\caption{TP--EP baseline with correlated placement. Each token stays on the rank of its canonical sequence slice, so its rows still cross the network even when its experts share one GPU.}
\label{fig:shuffled_tp_ep_baseline}
\end{figure}

\begin{figure*}[!t]
    \centering
    \includegraphics[width=\textwidth]{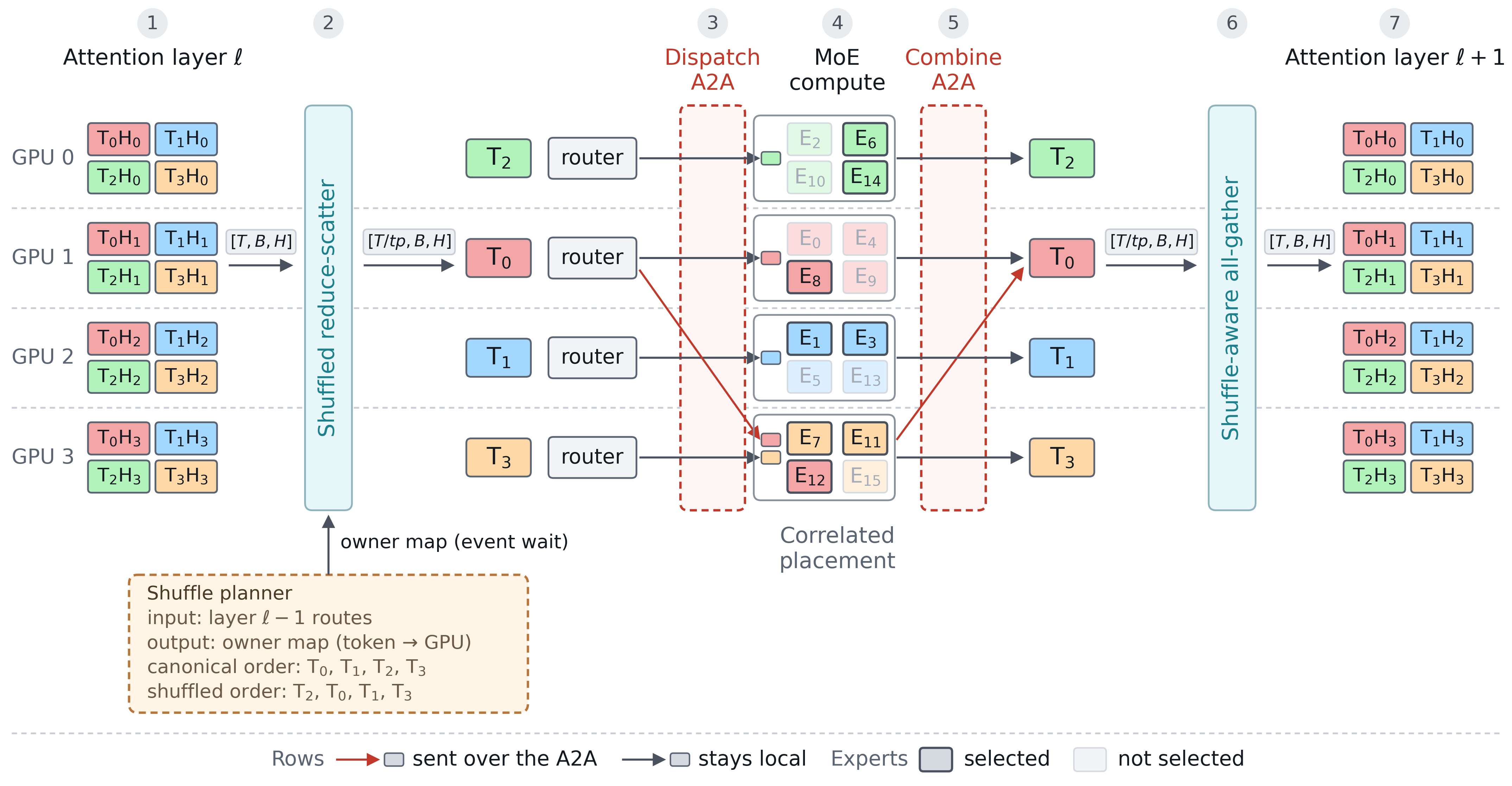}
    \caption{Token shuffling end to end. After the attention of layer $\ell$ \protect\stepnum{1}, the shuffle planner predicts each token's owner rank from its layer-$(\ell{-}1)$ routes, and the shuffled reduce-scatter \protect\stepnum{2} delivers each row to that rank. The dispatch \protect\stepnum{3}, the expert computation \protect\stepnum{4} and the combine \protect\stepnum{5} then find most of the token's experts local. The shuffle-aware all-gather \protect\stepnum{6} restores canonical order before the next attention block \protect\stepnum{7}.}
    \label{fig:shuffled_tp_ep_e2e}
\end{figure*}
Expert placement decides where experts live and does not change where tokens live. With TP--EP parallelism, the MoE layer runs on the sequence shard that the reduce-scatter after attention assigns to each rank. Each rank thus owns a fixed, contiguous slice of the sequence, and a token's rank is set by its position alone. Even when the experts a token needs have been placed together on one GPU, the token is actually on that GPU with probability only $1/\mathrm{EP}$. Thus, with probability $1-1/\mathrm{EP}$ the token's row must still cross the wire in the dispatch and return in the combine. Figure~\ref{fig:shuffled_tp_ep_baseline} shows this for four ranks under correlated placement. The conventional reduce-scatter places T0 on GPU0, T1 on GPU1 and so on, although T0's experts sit together on GPU1 and T1's on GPU2, so their rows still leave the owner's GPU.

The cross-layer correlation of Section~\ref{sec:affinity} (Figure~\ref{fig:across-affinity}) lets us move the token as well. The experts a token used at layer $\ell$ predict, well above chance, which GPU will host its experts at layer $\ell{+}1$, and this prediction is available a whole attention block before the layer-$(\ell{+}1)$ router runs. Token shuffling (Figure~\ref{fig:shuffled_tp_ep_e2e}) uses this window. We call the GPU that holds a token in the MoE layer its \emph{owner}. The reduce-scatter that follows attention first permutes tokens so that each goes to its predicted owner and then scatters them, so the following MoE dispatch finds that most of the token's experts are now local. The rest of this section describes where ownership can change without extra communication, how the owner is predicted, and how the planner is hidden behind attention.

\paragraph{Changing Ownership Inside Existing Collectives}
The token-shuffling construction requires the TP group and the EP group to be the same set of ranks ($\mathrm{TP}=\mathrm{EP}$). In TP--EP, a reduce-scatter follows the attention output projection and returns the sequence to its sharded layout. An all-gather precedes the next block's QKV projection and reassembles the full sequence on every rank (Figure~\ref{fig:shuffled_tp_ep_e2e}). Both collectives already move every token between ranks. Which rank receives which token in the reduce-scatter is set by indexing, so changing it does not change the number of bytes sent. We therefore fuse a permutation into each collective. The fused reduce-scatter sends every token to its planned owner instead of the rank that holds it in the default contiguous sequence layout. It also moves the token's row of the residual stream, so the residual addition happens on the new owner. The fused all-gather restores the original token order on the way back, because attention depends on token order. The MoE layer sees a local slice whose tokens mostly have their experts on the same GPU, and the attention layer is unchanged. Token shuffling thus adds no additional collective. Its two fused permutations cost a few milliseconds per step, compared with the hundreds of milliseconds removed from the all-to-all due to decreased communication (Section~\ref{sec:results}).

\paragraph{Predicting the Owner}
The shuffle planner answers one question per token: given the experts it just
visited, on which rank will its next experts most likely live? Routing is
strongly correlated across layers, so the pair of experts a token used at
the two preceding MoE layers is a good predictor of where it lands next.
Offline, from routing traces and the chosen placement, we count for every
MoE layer how often each pair sends the token to each rank; the fewer
observations a pair has, the more weight we give to the single preceding
expert instead. At runtime the previous layers' routes are the only input, and
the planner adds no collective. The lookup gives each token a preferred
owner and a confidence score. Sequence parallelism requires every rank to own
exactly $T/\mathrm{EP}$ tokens, so tokens are assigned in decreasing
confidence, each to its most preferred rank with capacity left. We give
the full algorithm in Algorithm~\ref{alg:affinity-shuffle}.

\begin{figure}[!h]
  \centering
  \includegraphics[width=0.95\columnwidth]{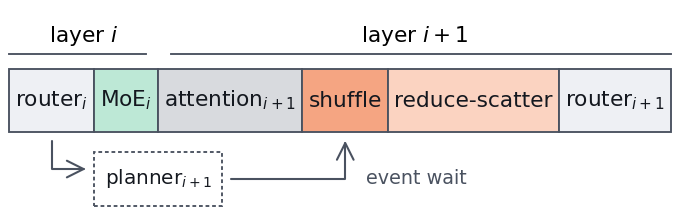}
  \caption{The planner runs on a side stream between the layer-$i$ routes
  and the layer-$(i{+}1)$ reduce-scatter.}
  \label{fig:planner}
\end{figure}

\paragraph{Hiding the Shuffle Planner} \label{para:shuffle-planner}
Prediction and quota assignment cost about a third of the forward
all-to-all time at TP16~EP16 so, if executed sequentially, this would forfeit most of the
gain. However, crucially, the plan depends only on routes up to layer $i$ and is needed only at
the layer-$(i{+}1)$ reduce-scatter, so we can run the planner on a side
stream, concurrently with the expert computation and the next attention block, and join it with a
single event wait (Figure~\ref{fig:planner}). In practice the main stream waits about one millisecond per step for the plan (0.02\,\% of the step), plus one fused permute-plus-residual kernel.


\section{Experimental Results}
\label{sec:results}

All experiments use the two models of Section~\ref{sec:mot-a2a}. Both are 12-layer MoE models with $d_{\mathrm{model}}=4096$ and 128 experts of FFN width 1024 per layer, one with top-2 and one with top-6 routing. Both use a sequence-level auxiliary loss with coefficient $10^{-2}$ for load balancing. We train each for 2.1 billion tokens (2000 steps) on FineWeb-Edu~\citep{fineweb} with a sequence length of 8192 and a global batch size of 128. 

We take all measurements on a cluster of nodes with eight AMD Instinct MI300X GPUs each, fully connected within a node by xGMI (one direct link between each pair of GPUs) and connected across nodes by eight 100\,Gb/s RoCE NICs per node, one per GPU. All timing runs load the step-2000 checkpoint of the respective model and run on held-out data from the same corpus, so the routing decisions are those of a trained model. We run each experiment twice. The first run consists of 14 clean steps; we report the median over the last 10 steps, after discarding the first 4 as warm-up (checkpoint loading and communicator initialization). The second run consists of 14 steps with the region profiler and all-to-all metrics enabled, which provide the time and traffic breakdowns, following the same procedure. Speedups are relative to the baseline, which is the Megatron-LM all-to-all dispatcher with contiguous placement. We compare three configurations against it. These are deduplication, deduplication with correlated expert placement, and, on layouts with $\mathrm{TP}=\mathrm{EP}$, token shuffling on top of both. The figures label the baseline and the three configurations B, D, P and S. We implement all methods in Megatron-LM~\citep{megatron-lm}. The appendix repeats the top-6 placement evaluation with two other load-balancing methods (Figure~\ref{fig:ablation_routers}). The correlations are present under all three, with different strengths.

\subsection{Building Expert Correlation Within and Across Layers}
\label{sec:res-corr}

Correlated placement and the token-shuffling predictor both use correlation tables fitted offline from recorded routing. We measure how many tokens are needed for these estimates to stabilize and how quickly they go stale during training. We test both cases on held-out data.

\paragraph{A Few Thousand Tokens Suffice}

To measure how much routing data the tables need, we fit them on different numbers of tokens from iterations 1983--1995 and evaluate them on held-out iteration 1999. With tables fitted on 1{,}024 tokens, correlated placement removes 23.6\,\% of dispatched rows at top-2 EP8 and 57.9\,\% at top-6. With 4{,}096 tokens, the Pearson correlation between the fitted and the held-out co-selection matrices is 0.93 for top-2 and 0.98 for top-6, and the placement and prediction results are within one percentage point of those obtained with all 524k tokens. A single microbatch from one rank is therefore enough to fit both tables. The tables capture a property of the trained router and do not depend on the particular tokens used to fit them. This implies that the correlation structure is robustly discoverable from relatively few tokens.

\paragraph{The Structure Forms Early and Remains Stable}
To measure how early the correlations form, we fit the tables on routing from a single training step and evaluate them on routing from step 1999, the end of training. Tables fitted at step 199 already give about half of the final benefit. At top-2 EP8, the placement built from them removes 14.4\,\% of dispatched rows, compared with 26.5\,\% for tables fitted at the end of training. At top-6 the corresponding numbers are 42.7\,\% and 57.7\,\%. Tables fitted at step 1000 come within 2--3 percentage points of the final values, and tables fitted at step 1400 or later give nearly the same results as the final tables. The correlation between the fitted and the final co-selection matrices follows the same pattern, rising from 0.4--0.5 at step 199 to above 0.93 from step 1400 on. The predictor improves at the same rate. At top-2 EP8, its hit rate is 12.7\,\% for tables fitted before the routing correlations have formed, close to the chance level of 12.5\,\%. It rises to 54\,\% for tables fitted at step 1000 and 57\,\% for tables fitted at step 1800. For our 2000-step runs, fitting the tables once in the second half of training is therefore enough. For longer runs, we expect refitting about every 1000 steps to be sufficient, and that each refit needs only one microbatch of recorded routes.

\subsection{Correlated Expert Placement}
\label{sec:res-placement}

\paragraph{All-to-All Communication Volume}
Correlated expert placement accounts for most of the reduction in all-to-all traffic. Under contiguous placement, deduplication removes only the duplicates that land on the same rank by chance. This removes 6\,\% of rows at top-2 and 25\,\% at top-6 for EP8, and 3\,\% and 12\,\% at EP16. With correlated placement, the communication volume falls by 1.36$\times$ at top-2 EP8 and 2.36$\times$ at top-6 EP8 relative to the baseline. The reduction grows with top-$k$, since more of a token's experts can share a rank. The effect holds as EP widens and each rank holds fewer experts, at 1.29$\times$/1.22$\times$/1.12$\times$ for top-2 and 1.95$\times$/1.54$\times$/1.24$\times$ for top-6 at EP16/32/64. Dispatch and combine carry identical byte counts by construction, since every row that goes out comes back, so the byte reduction applies identically to the backward pass also.

\begin{figure*}[!ht]
    \centering
    \includegraphics[width=0.8\linewidth]{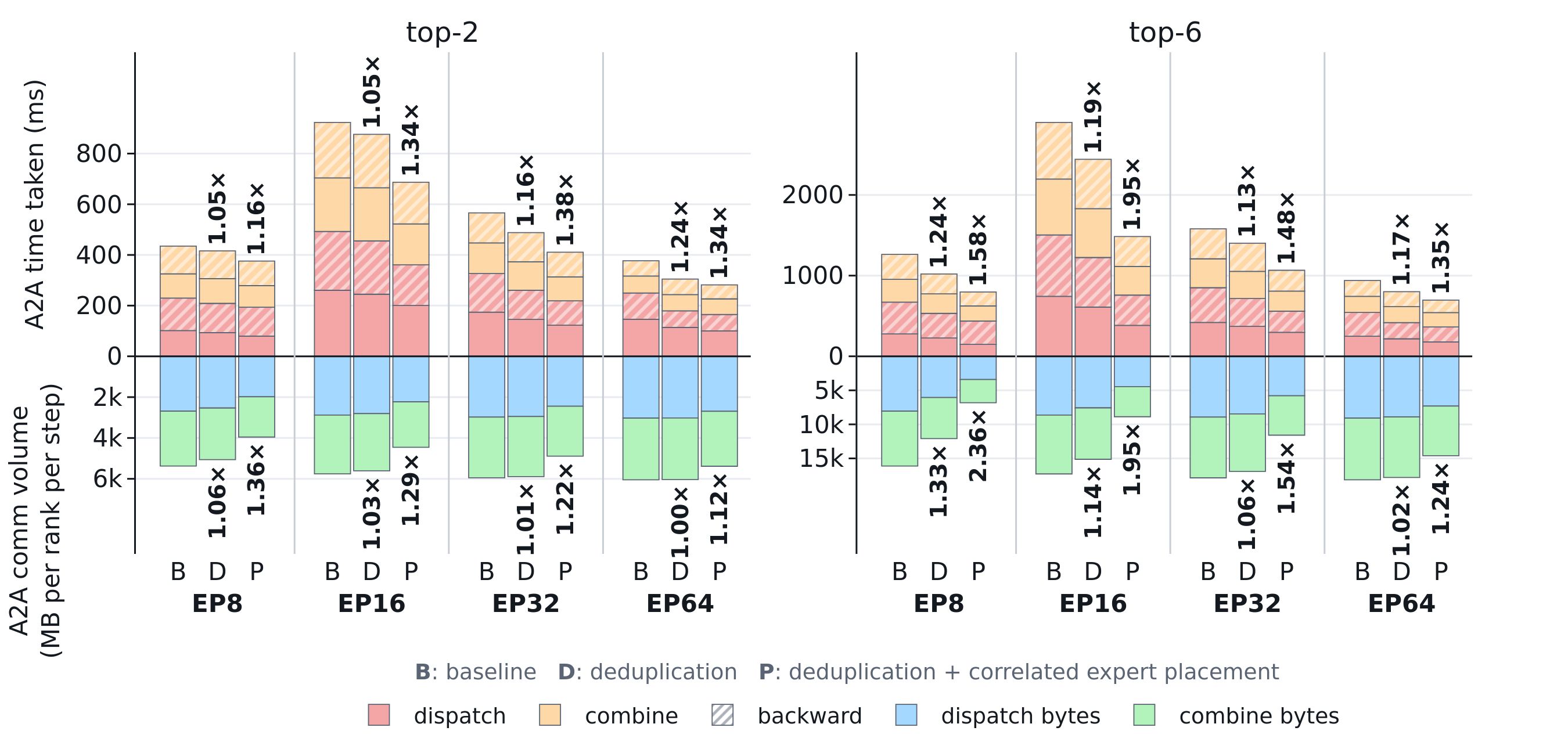}
    \caption{All-to-All Cost Under Expert Parallelism. A2A time (top) and communication volume per rank (bottom) with EP8--EP64, for top-2 and top-6 routing. B is the baseline, D adds deduplication, and P adds correlated placement to deduplication.}
    \label{fig:placement_a2a}
\end{figure*}

\paragraph{All-to-All Communication Time}
Across nodes, A2A time falls by 1.34$\times$/1.38$\times$/1.34$\times$ for top-2 and
1.95$\times$/1.48$\times$/1.35$\times$ for top-6 at EP16/32/64. Within a node (EP8), where
the collective runs over xGMI, it falls by 1.16$\times$ at top-2 and 1.58$\times$ at top-6.
The reduction appears in all four collectives, dispatch and combine in both the forward and
the backward pass.

\paragraph{End-to-End Step Time}

\begin{figure*}[!h]
    \centering
    \includegraphics[width=0.8\linewidth]{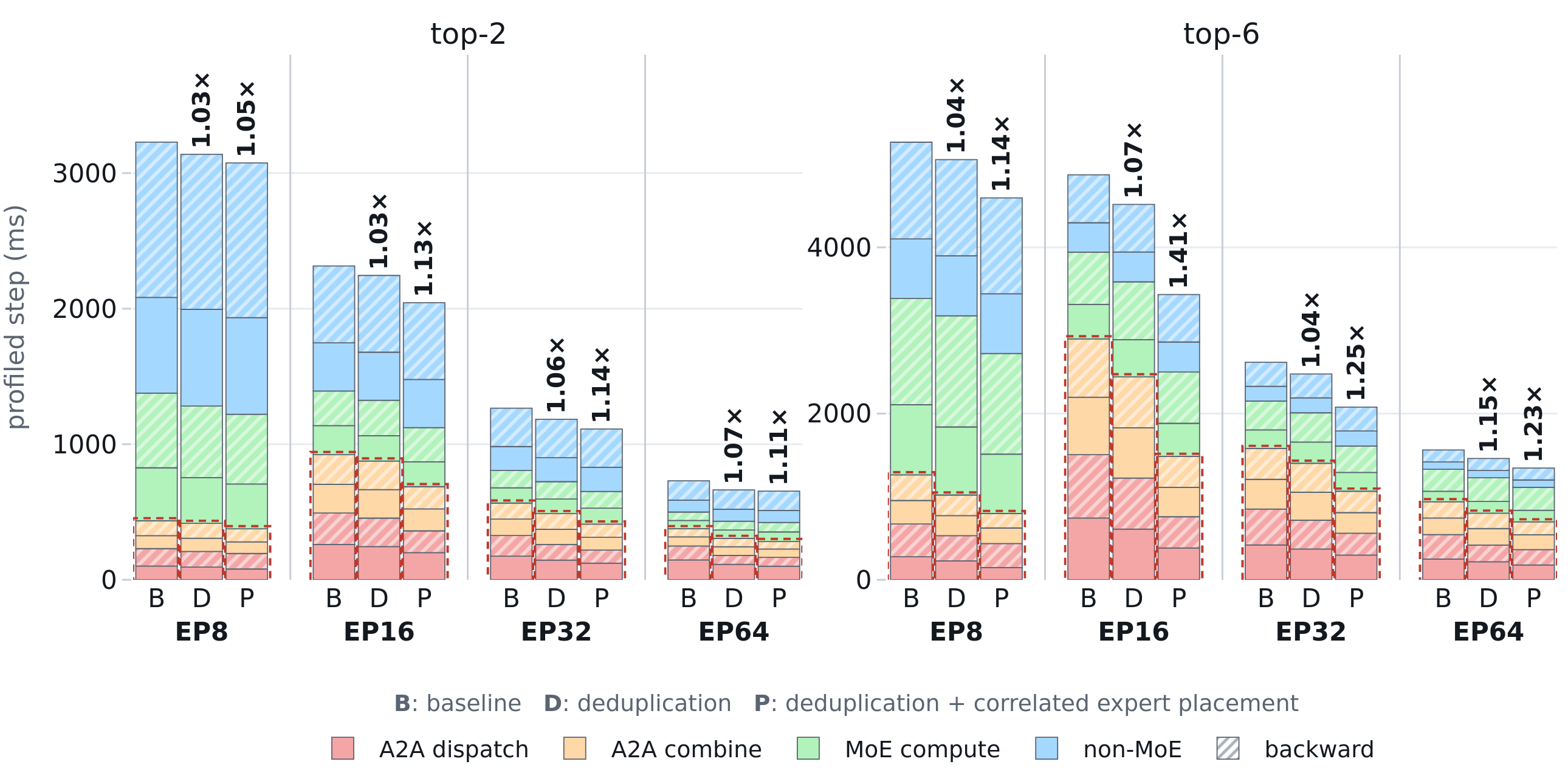}
    \caption{Step-time breakdown at EP8–EP64 for top-2 and top-6 routing; dashed boxes indicate A2A share.}
    \label{fig:e2e_placement}
\end{figure*}
Non-MoE computation takes nearly the same time in all configurations (blue in
Figure~\ref{fig:e2e_placement}). The step speedup therefore depends on how much A2A time is
saved and on how large a share of the step the A2A takes. At top-2 the A2A share is
13--52\,\%, and correlated placement speeds up the step by 1.05$\times$ at EP8, 1.13$\times$
at EP16, 1.14$\times$ at EP32 and 1.11$\times$ at EP64. At top-6 the share is 24--60\,\%, and
the speedups are 1.14$\times$ at EP8, 1.41$\times$ at EP16, 1.25$\times$ at EP32 and
1.23$\times$ at EP64. The speedup is larger at top-6 and is largest when the A2A crosses node
boundaries.
\subsection{Token Shuffling}
\label{sec:res-shuffle}

\begin{figure*}[!t]
    \centering
    \includegraphics[width=0.8\linewidth]{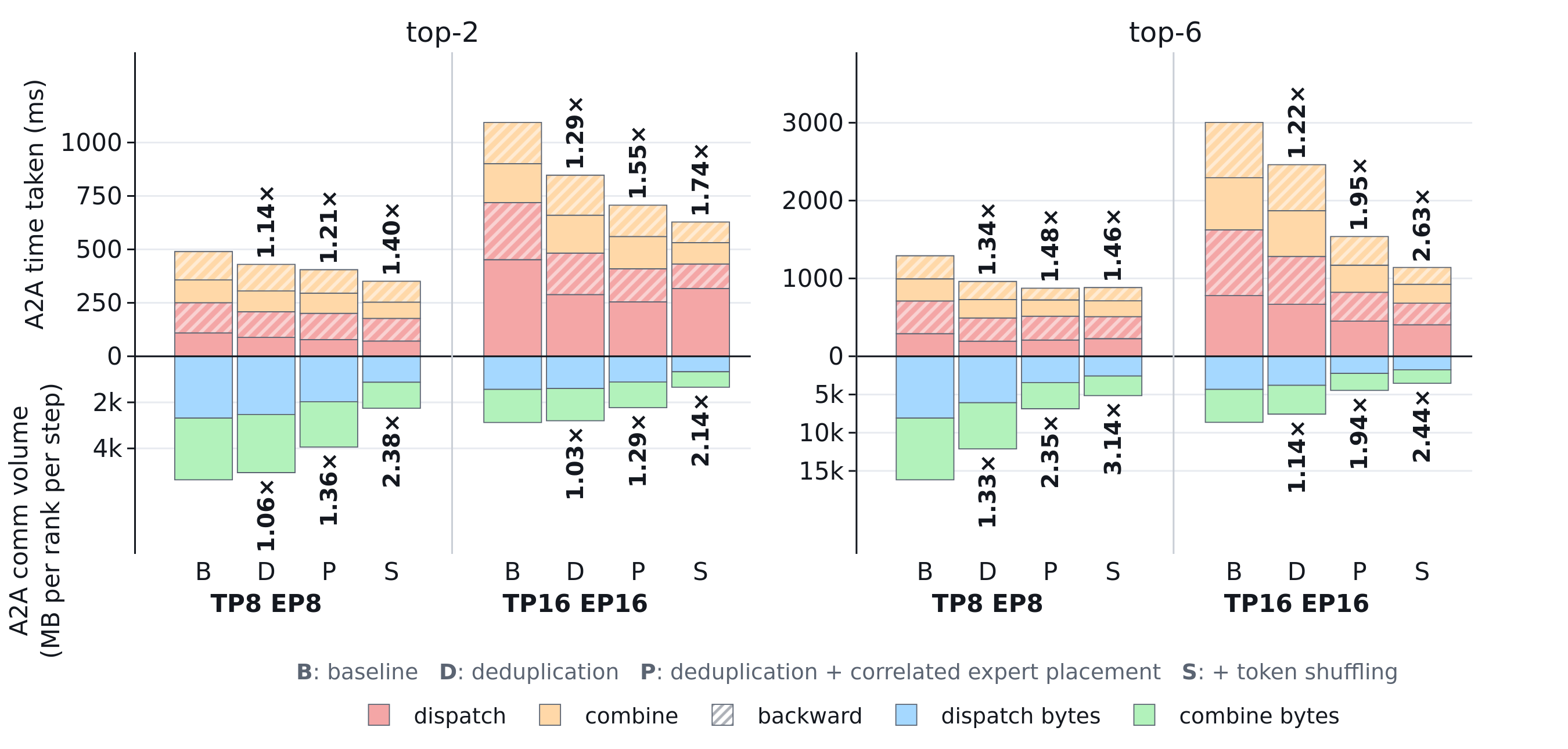}
    \caption{All-to-All Cost on the $\mathrm{TP}=\mathrm{EP}$ Layouts. Same encoding as Figure~\ref{fig:placement_a2a}, with token shuffling as the fourth configuration (S). The shuffle places tokens on the GPUs that hold their experts, so more of the dispatch stays on the device and less traffic crosses the network.}
    \label{fig:a2a_shuffled}
\end{figure*}

\begin{figure*}[!t]
    \centering
    \includegraphics[width=0.8\linewidth]{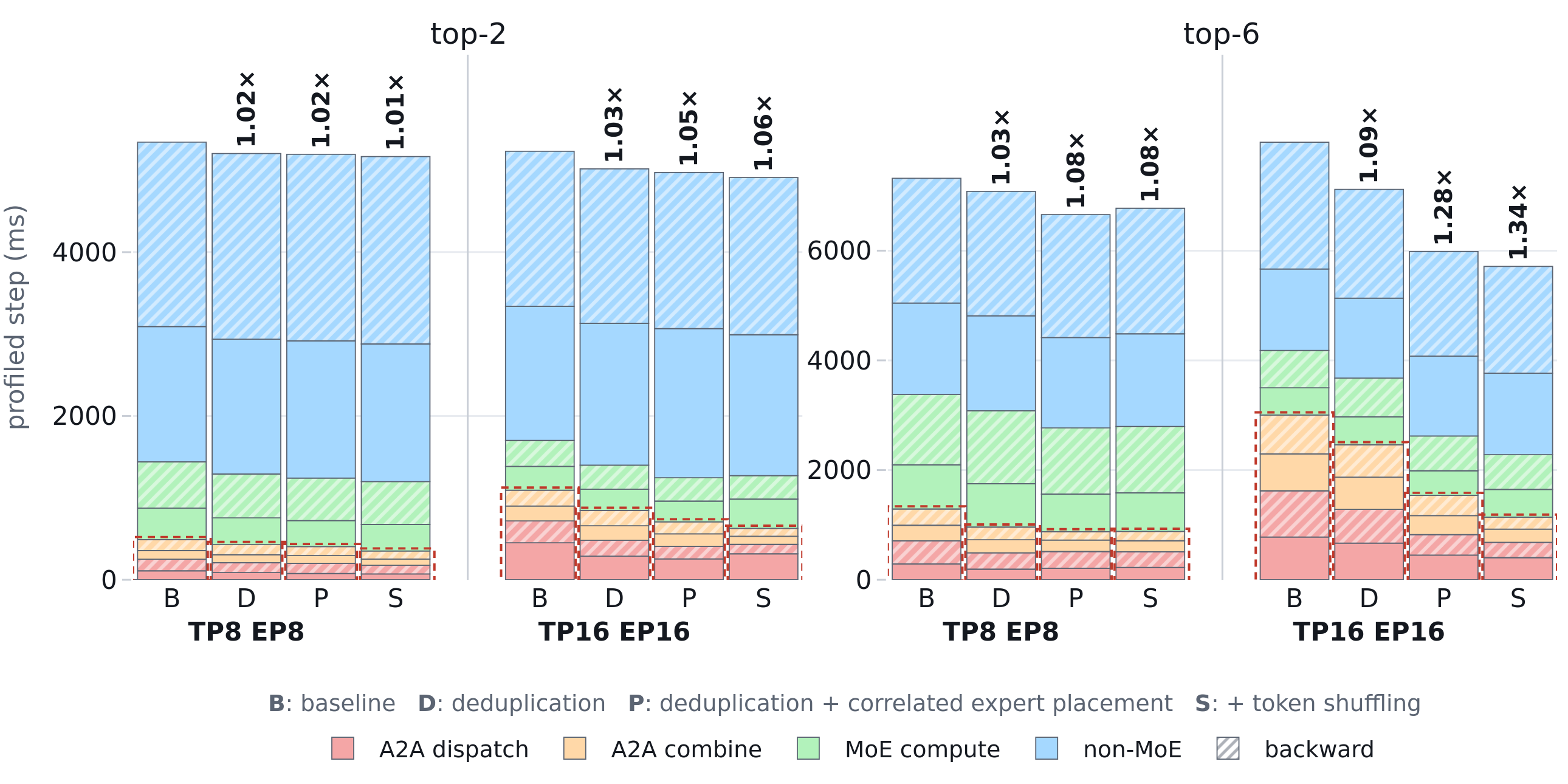}
    \caption{Step-time breakdown on the $\mathrm{TP}=\mathrm{EP}$ layouts (TP8\,EP8, TP16\,EP16) with token shuffling
  added as S. Same encoding as Figure~\ref{fig:e2e_placement}.}
    \label{fig:fig_shuffle_step_breakdown}
\end{figure*}

Token shuffling applies where the TP group and the EP group coincide. We evaluate it at TP8~EP8 on one node and TP16~EP16 across two nodes. Figure~\ref{fig:a2a_shuffled} shows its all-to-all results alongside deduplication and correlated placement, and Figure~\ref{fig:fig_shuffle_step_breakdown} shows the end-to-end step time.

\paragraph{Token–Expert Locality}

Without token shuffling, a token's expert slots are on its own GPU with probability $1/\mathrm{EP}$, which is 12.5\,\% at EP8 and 6.3\,\% at EP16. Token shuffling raises this to 59\,\% and 53\,\% for top-2 and to 57\,\% and 46\,\% for top-6, close to the offline hit rates in Table~\ref{tab:placement}. The dispatcher still counts these rows, but they stay on the GPU and never cross the network. Relative to the baseline, wire bytes fall by 2.4$\times$ at EP8 and 2.1$\times$ at EP16 for top-2, and by 3.1$\times$ and 2.4$\times$ for top-6. With placement alone, the reductions are 1.4$\times$ and 1.3$\times$ for top-2 and 2.4$\times$ and 1.9$\times$ for top-6.
\paragraph{All-to-All Communication Time}

At TP16~EP16, token shuffling raises the A2A reduction from 1.55$\times$ (placement alone) to 1.74$\times$ at top-2 and from 1.95$\times$ to 2.63$\times$ at top-6. On a single node (TP8~EP8), token shuffling reduces bytes by a similar or larger factor, but reduces the time cost by a smaller factor. The A2A reductions are 1.40$\times$ at top-2 and 1.46$\times$ at top-6. At top-6 on one node, the A2A reduction with token shuffling is slightly below that of placement alone (1.48$\times$). On xGMI, the collective time is set by the busiest GPU pair and does not follow the aggregate bytes. Concentrating a token's rows on one owner raises the traffic on the busiest link even as it lowers the total. The two fused permutations of token shuffling add a few milliseconds per step. The planner runs behind attention, and the main stream waits about one millisecond per step for it (Section~\ref{sec:shuffle}).

\paragraph{End-to-End Step Time} On the TP\,=\,EP layouts (Figure~\ref{fig:fig_shuffle_step_breakdown}) the tensor-parallel
collectives inflate the non-MoE part, so the A2A is a smaller share of the step: 9--21\,\% at
top-2 and 18--38\,\% at top-6. Token shuffling cuts the A2A time itself by 1.40$\times$ and
1.74$\times$ at top-2 and by 1.46$\times$ and 2.63$\times$ at top-6 (TP8\,EP8 and TP16\,EP16),
but the step speedup is limited by how much of the step the A2A takes. At top-2 it stays within 1.01--1.06$\times$; at TP16\,EP16 it grows from 1.03$\times$ with deduplication to 1.05$\times$ with correlated placement and 1.06$\times$ with token shuffling. At top-6 the gains are 1.08$\times$ at TP8\,EP8 and
1.34$\times$ at TP16\,EP16, where shuffling improves the speedup from 1.28$\times$ to 1.34$\times$.


\section{Conclusion}
We showed that expert routing in MoE pretraining is correlated both within a layer and across layers, and that these correlations form within about one billion tokens and then stay stable. Correlated expert placement uses the within-layer correlation to put co-selected experts on the same GPU, and a deduplicating dispatcher sends each token at most once to its destination GPU. Token shuffling uses the cross-layer correlation to move each token to the GPU predicted to hold its next experts, subject to per-rank capacity constraints. Moreover, this shuffle occurs inside the reduce-scatter that sequence parallelism already performs, leading to almost no overhead. Neither method changes routing decisions or expert parameters. On a cluster of nodes with eight MI300X GPUs each, the two methods reduce all-to-all time by 1.16--2.63$\times$ and end-to-end step time by up to 1.41$\times$, with the largest gains at top-6 when the EP group spans several nodes. Our evaluation covers one 12-layer model with top-2 and top-6 routing, trained for 2.1B tokens. We believe the expert correlation holds across model families and scales, and leave other parallelism layouts, such as context parallelism, and EP inference to future work.









\bibliographystyle{plainnat}
\bibliography{main}

\clearpage
\clearpage
\onecolumn
\appendix

\section{Supplementary Material}
\label{app:supplementary}

\subsection{Experimental Setup}
\subsubsection{Model Architecture}
Table~\ref{tab:p2_arch} summarizes the architecture and pre-training
configuration of the model used throughout our experiments. Unless stated
otherwise, every result uses this configuration, including its router
(pre-softmax top-$k$ with $k=2$ or $k=6$ and a sequence-level auxiliary loss of $10^{-2}$). The
router ablation in Section~\ref{sec:router_ablation} keeps the architecture
fixed and varies only the routing recipe: the pre-softmax router with auxiliary
loss $10^{-2}$ and $10^{-1}$, and the auxiliary-loss-free router of
DeepSeek-V3~\citep{deepseekv3} (sigmoid scores with a per-expert bias updated at rate $10^{-3}$
and a $10^{-4}$ sequence-level loss).

\begin{table}[!h]
  \centering
  \small
  \begin{tabular}{lll}
    \toprule
    & \textbf{Symbol} & \textbf{Value} \\
    \midrule
    \multicolumn{3}{l}{\textbf{Transformer}} \\
    Layers                       & $L$                     & 12, every layer an MoE layer \\
    Model width                  & $d_{\text{model}}$      & 4096 \\
    Attention heads              & $n_{\text{head}}$       & 16 \\
    Head dimension               & $d_{\text{head}}$       & 32 \\
    Attention width              & $n_{\text{head}}\, d_{\text{head}}$ & 512 \\
    Normalization                & --                      & pre-LayerNorm, $\epsilon = 10^{-5}$ \\
    Positional encoding          & --                      & learned absolute \\
    Vocabulary                   & $|V|$                   & 151{,}669 (Qwen3 tokenizer; padded to 151{,}680) \\
    Biases / dropout             & --                      & none / 0 \\
    Embeddings                   & --                      & input and output tied \\
    \midrule
    \multicolumn{3}{l}{\textbf{Mixture of experts}} \\
    Experts per layer            & $E$                     & 128, no shared expert \\
    Experts per token            & $k$                     & 2 or 6 \\
    Expert hidden size           & $d_{\text{ff}}$         & 1024 ($d_{\text{model}} \to d_{\text{ff}} \to d_{\text{model}}$, GELU) \\
    Routing                      & --                      & top-$k$ of $E$, dropless \\
    Router                       & --                      & FP32 linear gate; softmax over all $E$, then top-$k$; \\
                                 &                         & gate weights = selected softmax probabilities \\
    Load-balancing loss          & $\alpha_{\text{aux}}$   & sequence-level auxiliary loss, $10^{-2}$ \\
    Dispatch                     & --                      & all-to-all, grouped GEMM \\
    \midrule
    \multicolumn{3}{l}{\textbf{Size}} \\
    Total parameters             & $N$                     & 13.6\,B (13.0\,B transformer, 0.6\,B embeddings) \\
    Expert parameters            & $N_{\text{exp}}$        & 12.9\,B (1.07\,B per layer) \\
    Active per token             & $N_{\text{act}}$        & $\approx 0.31$\,B at top-2, $\approx 0.71$\,B at top-6 (non-embedding) \\
    \midrule
    \multicolumn{3}{l}{\textbf{Pre-training}} \\
    Data                         & --                      & FineWeb-Edu, 98/2 train/validation split \\
    Sequence length              & $S$                     & 8192 \\
    Global batch                 & $B$                     & 128 sequences ($B \cdot S \approx 1.05$\,M tokens per step) \\
    Steps / tokens               & $T$                     & 2000 / $\approx 2.1$\,B tokens \\
    Optimizer                    & --                      & AdamW, $\beta_1 = 0.9$, $\beta_2 = 0.95$, weight decay 0.1, gradient clip 1.0 \\
    Learning rate                & $\eta$                  & $10^{-5}$ peak, 200 warm-up steps, cosine decay toward $10^{-6}$ \\
    Initialization               & $\sigma_{\text{init}}$  & 0.008 (normal) \\
    Precision                    & --                      & BF16; FP32 router and optimizer state \\
    Parallelism                  & TP / EP                 & 1 / 8, one 8-GPU node, micro-batch 2 \\
    \bottomrule
  \end{tabular}
  \caption{Model and pre-training configuration used in all experiments.}
  \label{tab:p2_arch}
\end{table}

\subsubsection{Training Framework} 
We use Megatron-LM~\citep{megatron-lm} as our training framework and implement our methodology (Section~\ref{sec:implementation}) on top of it, leaving the router, expert computation, optimizer, and training loop unchanged. Specifically, our implementation modifies three components. (i)~\emph{Token dispatcher}: We add a deduplicating variant of the all-to-all dispatcher that sends one row per unique (token, destination rank) pair instead of one per (token, expert) pair. Expert IDs and gate weights are carried with the activation, and the receiver expands the rows into expert-major order. The combine direction applies the same deduplication in reverse.

(ii)~\emph{Expert placement}: We extend the dispatcher's contiguous expert-to-rank assignment to support an arbitrary per-layer permutation. The permutation is applied to the routing map, and checkpoint expert weights are keyed by global expert ID, allowing the same checkpoint to be loaded with a different expert placement.

(iii)~\emph{Token shuffling}: For TP$=$EP layouts, we modify attention's sequence-parallel reduce-scatter to assign each token to an owner rank based on the second-order affinity table. Canonical token order is restored around attention, the output layer, and the loss. The route IDs the planner needs are included in the dispatcher's existing metadata all-gather, requiring no additional collective.

Routing decisions used to build the offline expert placement and affinity tables are recorded through a capture hook in the router. All variants train the same model and produce the same loss, differing only in how tokens are placed and moved across ranks.

\subsection{Ablation Over Routing Strategies}
\label{sec:router_ablation}

\paragraph{Load-Balancing Methods}
Our methods depend on the router producing correlated expert choices. To test how the
load-balancing method affects this, we trained two more top-6 models that differ from the
main top-6 model only in the balancing method. The first uses a sequence-level auxiliary loss
ten times stronger than the main model's (coefficient $10^{-1}$ instead of $10^{-2}$). It
gives the most balanced expert load of the three, with a maximum-to-mean load ratio of 1.20. The
second uses the auxiliary-loss-free method of DeepSeek-V3~\citep{deepseekv3}, which adds a
per-expert bias to a sigmoid router. Its load ratio is 1.37, and it reaches the lowest loss
of the three (4.39, against 4.46 for the main model and 4.64 for the stronger auxiliary
loss). We evaluate the same configurations on all three models.

\begin{figure}[!h]
  \centering
  \includegraphics[width=\linewidth]{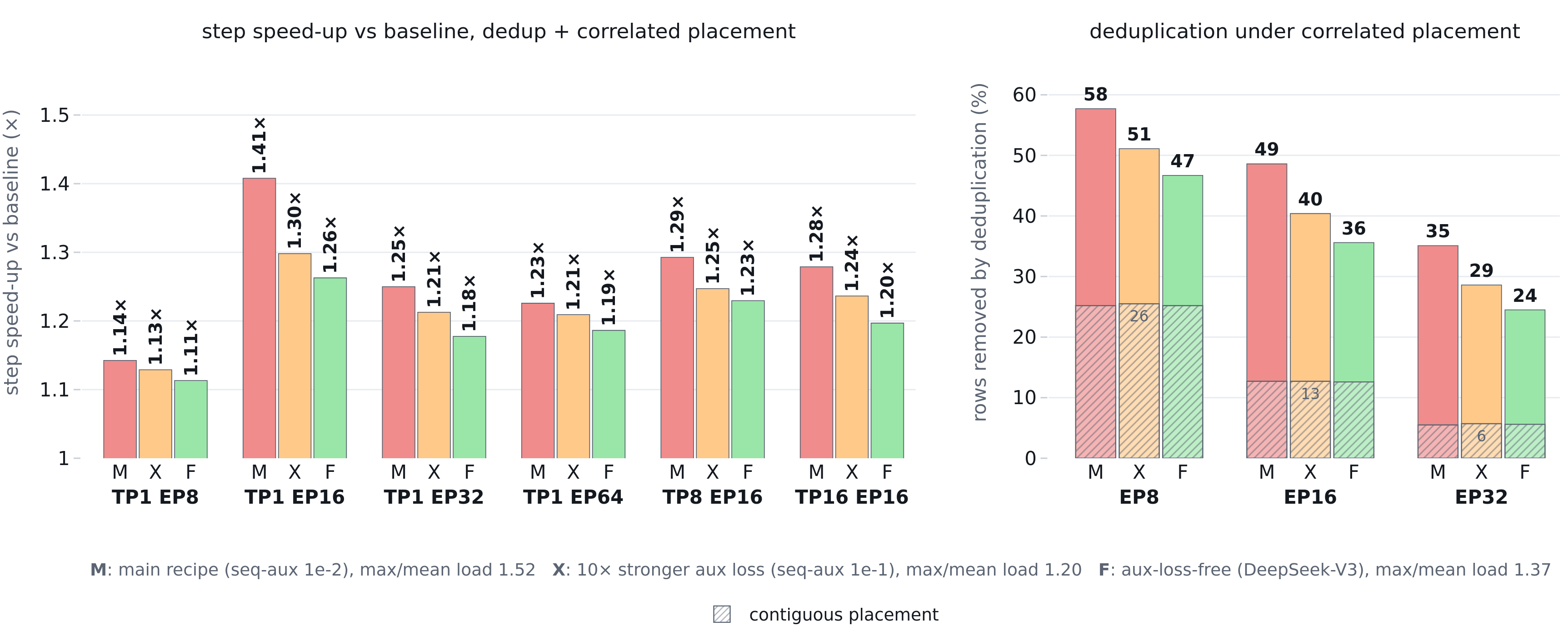}
  \caption{Router ablation on three top-6 models that differ only in load balancing: the main
  recipe (sequence auxiliary loss $10^{-2}$), a $10\times$ stronger auxiliary loss ($10^{-1}$),
  and the DeepSeek-V3 auxiliary-loss-free router. Left: step speedup of deduplication +
  correlated placement over that model's baseline, per layout. Right: the structure each router
  leaves to exploit, as the deduplication rate under correlated placement.}
  \label{fig:ablation_routers}
\end{figure}

\paragraph{Gains Follow Each Router's Co-Selection Structure}
Correlated placement remains effective across all three routing strategies.
With correlated placement, deduplication removes 58\,\%, 51\,\%, and 47\,\%
of dispatched rows at EP8 for the main model, the stronger auxiliary-loss
router, and the auxiliary-loss-free router, respectively, and 49\,\%, 40\,\%,
and 36\,\% at EP16. In comparison, contiguous placement removes only about 25\,\%
at EP8 and 13\,\% at EP16 for every router
(Figure~\ref{fig:ablation_routers}, right). These reductions translate into
consistent end-to-end gains across routing strategies: at TP1 EP16, step
speedups are 1.41$\times$, 1.30$\times$, and 1.26$\times$, respectively,
and remain at least 1.20$\times$ on every EP16 layout
(Figure~\ref{fig:ablation_routers}, left). Thus, the benefit of correlated
placement does not depend on a particular routing strategy; its magnitude
instead follows the co-selection structure produced by each router.

Load balance affects the realized communication time because an all-to-all
must wait for the most heavily loaded rank. At EP16, for example, the main
model removes 49\,\% of dispatched rows compared with 40\,\% for the stronger
auxiliary-loss router, but its speedup advantage is smaller
(1.41$\times$ versus 1.30$\times$), consistent with its higher load ratio
(1.52 versus 1.20). Correlated placement and deduplication therefore target
communication volume through expert co-selection, rather than communication
skew. Methods that directly reduce skew, such as relocating or replicating heavily loaded experts \citep{smartmoe, deepseekv3,moonep,gracemoe}, address a separate source of all-to-all cost and are complementary
to our approach.

\newpage 
\clearpage

\subsection{Per-Step Measurements}
\begin{figure}[!h]
  \centering
  \includegraphics[width=\linewidth]{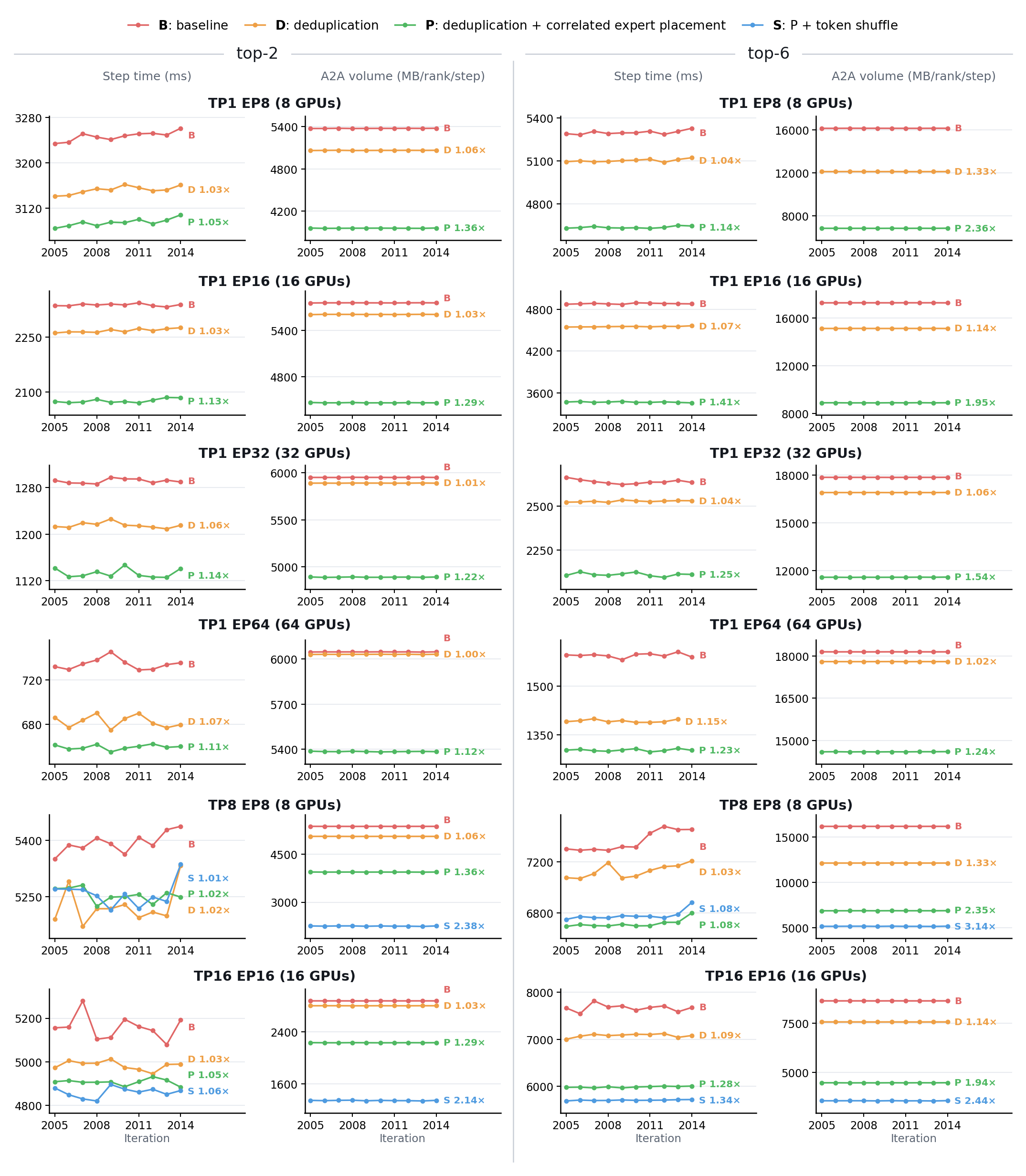}
  \caption{Step time (left) and A2A volume per rank (right) for the ten measured steps of each configuration, after the four warm-up steps. Data point labels give the ratio of the baseline median to the median of that configuration.}
  \label{fig:iter_time}
\end{figure}

Figure~\ref{fig:iter_time} shows the ten measured steps behind every reported speedup.
\subsection{Algorithms}

We give algorithms for the two components described in
Sections~\ref{sec:within} and~\ref{sec:shuffle}, using the notation of
Table~\ref{tab:notation}.

\begin{table}[!h]
\centering
\small
\caption{Notation.}
\label{tab:notation}
\begin{tabular}{@{}lp{0.72\linewidth}@{}}
\toprule
Symbol & Meaning \\
\midrule
$t$, $T$ & a token; number of tokens in the global sequence \\
$E$, $k$ & experts per MoE layer; experts selected per token (top-$k$) \\
$P$, $N$ & EP ranks (in Alg.~\ref{alg:affinity-shuffle} also the TP size); nodes \\
$\ell$, $m$ & MoE layer index; MoE position whose token owners are being planned \\
$\mathcal{S}_\ell(t)$ & the $k$ experts the router selects for token $t$ at layer $\ell$ \\
$W[e,e']$ & number of tokens that select both $e$ and $e'$ (co-selection count) \\
$\pi_\ell(e)$ & EP rank that holds expert $e$ at layer $\ell$ \\
$\nu(e)$, $\rho_n(e)$ & node of expert $e$; its rank within node $n$ \\
$I_{\max}$ & maximum number of swaps per partition \\
$F_m[e,r]$ & fraction of position-$m$ expert slots on rank $r$ for tokens that selected $e$ at $m{-}1$ \\
$S_m[e',e,r]$ & as $F_m$, conditioned on $e'$ at $m{-}2$ and $e$ at $m{-}1$ \\
$n_m[e',e]$ & number of trace tokens behind $S_m[e',e,\cdot]$ (support) \\
$Q_m[\mathrm{tok},r]$ & as $F_m$, conditioned on the token id (optional) \\
$\beta$, $w_{\max}$ & back-off strength; maximum weight on a conditioned table \\
$d_t$ & predicted distribution of token $t$'s position-$m$ expert slots over ranks \\
$d_t^{(1)}$, $d_t^{(2)}$ & largest and second-largest entries of $d_t$ \\
$\mathrm{owner}(t)$ & rank that holds token $t$'s row at MoE position $m$ \\
\bottomrule
\end{tabular}
\end{table}

\paragraph{Correlated placement (Alg.~\ref{alg:correlative-placement})}
The placement is computed offline, once per model. For each MoE layer we count
co-selections $W$ on the recorded routes and partition the experts into equal-size
groups that maximize the within-group count. \textsc{Partition} starts from
contiguous groups and applies the best improving pairwise swap until none remains
or $I_{\max}$ swaps are made; since every move is a swap, each rank keeps exactly
$E/P$ experts. On $N>1$ nodes the partition is hierarchical: experts are first split
across nodes, which targets cross-node traffic, and then across the $P/N$ ranks of
each node.

\begin{algorithm}[!h]
\caption{Correlated placement}
\label{alg:correlative-placement}
\small
\begin{algorithmic}[1]
\State \textbf{Input:} routing sets $\{\mathcal{S}_{\ell}(t)\}$, experts $E$,
EP ranks $P$, nodes $N$, budget $I_{\max}$
\State \textbf{Output:} balanced placement $\pi_\ell$ per MoE layer
\For{each MoE layer $\ell$}
    \State $W[e,e'] \gets \sum_t \mathbf{1}[e,e'\in\mathcal{S}_{\ell}(t)]$
    \If{$N=1$}
        \State $\pi_\ell \gets \Call{Partition}{W,P,I_{\max}}$
    \Else
        \State $\nu \gets \Call{Partition}{W,N,I_{\max}}$
        \For{node $n=0,\ldots,N-1$}
            \State $\mathcal{M}_n \gets \{e:\nu(e)=n\}$
            \State $\rho_n \gets \Call{Partition}{W_{\mathcal{M}_n},\tfrac{P}{N},I_{\max}}$
            \State $\pi_\ell(e) \gets n\tfrac{P}{N}+\rho_n(e)\ \ \forall e\in\mathcal{M}_n$
        \EndFor
    \EndIf
\EndFor
\Statex
\Procedure{Partition}{$W,G,I_{\max}$}\Comment{balanced}
    \State Initialize $G$ contiguous groups of $|W|/G$ experts
    \For{$i=1,\ldots,I_{\max}$}
        \State $(a,b)\gets$ swap with the largest gain in intra-group $W$
        \State \textbf{if} the gain is not positive \textbf{then break}
        \State Swap $a$ and $b$
    \EndFor
    \State \Return expert-to-group assignment
\EndProcedure
\end{algorithmic}
\end{algorithm}

\paragraph{Token shuffling (Alg.~\ref{alg:affinity-shuffle})}
The prediction tables are fit offline on the same traces as the placement; the
planner runs at every MoE position during training. It first predicts, for each
token, how its position-$m$ expert slots will spread over ranks (lines 3--7): the
first- and second-order tables are mixed per expert pair, with more weight on the
second-order table the more trace tokens support it. It then assigns owners greedily,
most confident token first, under a per-rank cap of $T/P$ so the layout stays
balanced (lines 8--11). The planner runs on a side stream during attention, and the
main stream waits for it only at the attention output projection, where the
reduce-scatter writes each row to its owner.

\begin{algorithm}[!h]
\caption{Token shuffling ($\mathrm{TP}=\mathrm{EP}=P$)}
\label{alg:affinity-shuffle}
\small
\begin{algorithmic}[1]
\State \textbf{Offline:} for each MoE position $m$, tables $F_m$, $S_m$, supports $n_m$ (optionally $Q_m$) from routing traces under placement $\pi$; back-off $\beta$, cap $w_{\max}$.
\State \textbf{Input:} top-$k$ sets $\mathcal{S}_{m-1}(t)$, $\mathcal{S}_{m-2}(t)$ of all $T$ tokens, gathered with the dispatcher's metadata all-gather at MoE position $m{-}1$.
\Statex \textit{// The shuffle planner runs on a side stream, overlapped with the expert computation at $m{-}1$ and the attention that follows; see \ref{para:shuffle-planner}.}
\For{each token $t$} \Comment{predict owner-rank distribution}
    \State $d_t \gets \frac{1}{k^2}\sum_{e \in \mathcal{S}_{m-1}(t)}\sum_{e' \in \mathcal{S}_{m-2}(t)} \big[(1-w_{e'e})\,F_m[e,\cdot] + w_{e'e}\,S_m[e',e,\cdot]\big]$,
    \Statex \hspace{\algorithmicindent}\hspace{\algorithmicindent} where $w_{e'e}=\min\!\big(n_m[e',e]/(n_m[e',e]+\beta),\,w_{\max}\big)$
    \State Optionally mix in $Q_m[\mathrm{tok}(t),\cdot]$ with the same back-off
    \State $d_t \gets d_t / \sum_r d_t[r]$
\EndFor
\State $\mathrm{cap}[r] \gets T/P$ for all $r$
\For{each $t$ in decreasing order of $d_t^{(1)} - d_t^{(2)}$} \Comment{most confident first}
    \State $\mathrm{owner}(t) \gets \arg\max_{r:\,\mathrm{cap}[r]>0} d_t[r]$; \; $\mathrm{cap}[\mathrm{owner}(t)] \gets \mathrm{cap}[\mathrm{owner}(t)]-1$
\EndFor
\State Build the new token order, its inverse, and the residual-row indices
\Statex \textit{// Main stream, at the attention output projection}
\State Wait for the planner
\State Reduce-scatter the projected rows to $\mathrm{owner}(\cdot)$; permute and add the residual in one fused kernel
\State MoE position $m$ dispatches from this layout, which persists until the next plan
\end{algorithmic}
\end{algorithm}


\subsection{Author Contributions}
\label{app:contributions}

\paragraph{Radha Gulhane} Implemented the methods and carried out all experiments. Led initial ideation and exploration.

\paragraph{Quentin Anthony} Contributed to the initial ideation and shaping of the idea, helped write the paper, and provided substantial guidance throughout.

\paragraph{Beren Millidge} Participated in technical discussions and provided suggestions throughout the project, and helped write the paper.



\end{document}